\documentclass[runningheads]{llncs}
\usepackage{graphicx}
\usepackage{comment}
\usepackage{amsmath,amssymb} % define this before the line numbering.
\usepackage{color}

\usepackage{epsfig}

\usepackage{multirow}
\usepackage{braket}
\usepackage{makecell}
\usepackage{xcolor}
\usepackage{siunitx}
\usepackage{amssymb}
\usepackage{pifont}
\usepackage{overpic}
\usepackage{float}

\makeatletter
\def\eg{\emph{e.g.}}

\def\etal{\emph{et al.}}
\def\aka{\emph{a.k.a.}}
\makeatother

\begin{document}
% \renewcommand\thelinenumber{\color[rgb]{0.2,0.5,0.8}\normalfont\sffamily\scriptsize\arabic{linenumber}\color[rgb]{0,0,0}}
% \renewcommand\makeLineNumber {\hss\thelinenumber\ \hspace{6mm} \rlap{\hskip\textwidth\ \hspace{6.5mm}\thelinenumber}}
% \linenumbers
\pagestyle{headings}
\mainmatter

\title{BABO: Background Activation Black-Out\\ for Efficient Object Detection}

\makeatletter
% *, 2, 3, ...
\renewcommand*{\@fnsymbol}[1]{\ensuremath{\ifcase#1\or *\or \dagger\or \ddagger\or
    \mathsection\or \mathparagraph\or \|\or **\or \dagger\dagger
    \or \ddagger\ddagger \else\@ctrerr\fi}}
\makeatother

% CAMERA READY SUBMISSION
% \begin{comment}
\titlerunning{BABO for Efficient Object Detection}
\author{Byungseok Roh$^{\dagger}$\thanks{Corresponding author} \and
Han-Cheol Cho\thanks{Equal contribution} \and
Myung-Ho Ju$^{\dagger}$ \and
Soon Hyung Pyo}
\authorrunning{B Roh et al.}
% First names are abbreviated in the running head.
% If there are more than two authors, 'et al.' is used.
%
\institute{Intel \\
\email{\{peter.roh, han-cheol.cho, lake.ju, sean.pyo\}@intel.com}}
% \end{comment}
%******************
\maketitle

\begin{abstract}
Recent advances in deep learning have enabled complex real-world use cases comprised of multiple vision tasks and detection tasks are being shifted to the edge side as a pre-processing step of the entire workload. 
Since running a deep model on resource-constraint devices is challenging, techniques for efficient inference methods are demanded.
In this paper, we present an objectness-aware object detection method to reduce computational cost by sparsifying activation values on background regions where target objects don't exist.
Sparsified activation can be exploited to increase inference speed by software or hardware accelerated sparse convolution techniques.
To accomplish this goal, we incorporate a light-weight objectness mask generation (OMG) network in front of an object detection (OD) network so that it can zero out unnecessary background areas of an input image before being fed into the OD network.
In experiments, by switching background activation values to zero, the average number of zero values increases further from 36\% to 68\% on MobileNetV2-SSDLite even with ReLU activation while maintaining accuracy on MS-COCO.
This result indicates that the total MAC including both OMG and OD networks can be reduced to 62\% of the original OD model when only non-zero multiply-accumulate operations are considered.
Moreover, we show a similar tendency in heavy networks (VGG and RetinaNet) and an additional dataset (PASCAL VOC).
%The code will be released.
 % Objectness Mask, 
\end{abstract}

\section{Introduction}

In recent years, we have witnessed the dramatic advancement and the success of deep learning.
It has brought a great leap in the field of computer vision such as image classification \cite{ResNet_2016,AlexNet_2012,GoogLeNet_2015} and object detection \cite{SingleShotDetector,FasterRCNN_2015}.
The impact is not limited to the research society.
Industries are already using deep learning for the development of artificial intelligence (AI) softwares and services such as autonomous driving, surveillance system, virtual assistant and health-care.

Training and running a deep learning model requires a large amount of computation power.
It limits the scalability and applicability of deep learning, especially for resource-constraint environments.
To improve the performance in the deep neural networks, hardware designs for dealing with zero values \cite{hardware_2017}, handling sparsity and high-precision outliers \cite{hardware_2018,sbnet_2018}, on-chip convolutional neural network (CNN) models \cite{OnChipCNNCifar10_2018,HWAccEscher_2017} and specialized accelerators \cite{Movidius_SmartSensor_2018,Movidius_OD_2018} have been proposed.
In the perspective of software acceleration, it includes light-weight network design \cite{PVANET_2016,YOLO_2016,MobileNetV2_2018}, network pruning \cite{LotteryTicketHypothesis_2019,PruningWeightsAndConnections_2015} and network quantization \cite{BinaryConnect_2015,BinaryNet_2016}.
Our approach is mostly motivated by the previous works \cite{hardware_2017,sbnet_2018} exploiting the sparsity of weights and activation.

\begin{figure}[t]
\centering
\def\arraystretch{0.8}
\setlength{\tabcolsep}{0.2em}
\begin{tabular}{cccc}
\includegraphics[width=0.23\linewidth]{figures/figure01_input.png} &
\includegraphics[width=0.23\linewidth]{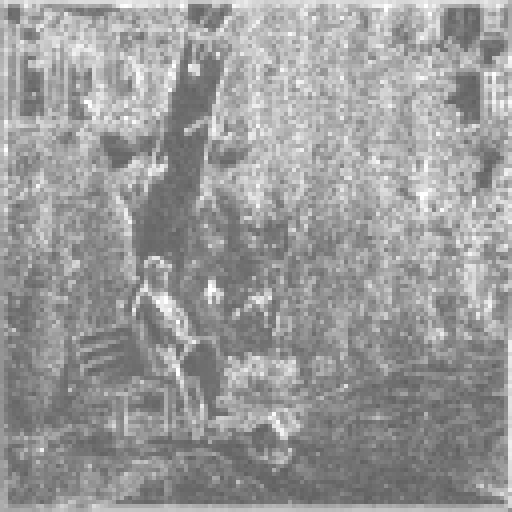} &
\includegraphics[width=0.23\linewidth]{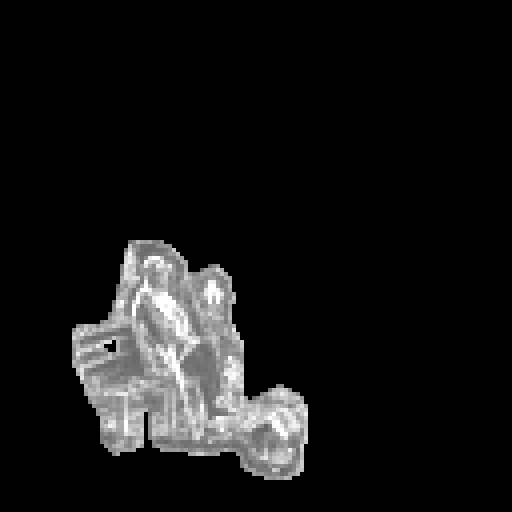} &
\includegraphics[width=0.23\linewidth]{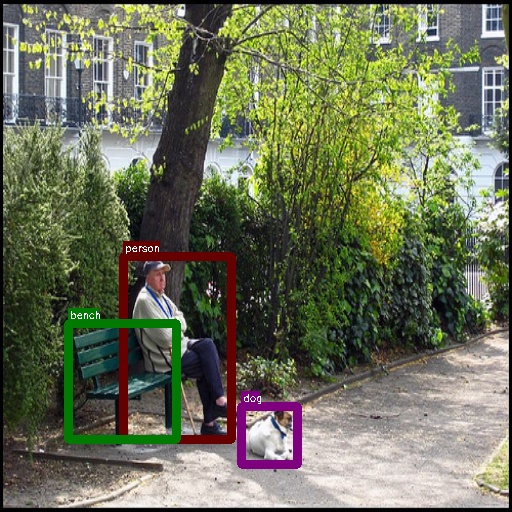}\\
{\footnotesize (a) input} & {\footnotesize (b) vanilla} & {\footnotesize (c) proposed} & {\footnotesize (d) our result}
\end{tabular}
\vspace{-1mm}
\caption{Comparison of generated feature maps; (a) input, (b) feature map generated by a modern object detection model, (c) the proposed model's feature map, (d) detection result of the proposed method.
While object boundaries are rough and some foreground pixels are lost, the masked feature map keeps most information of foreground regions.
}\vspace{-4mm}
\label{fig:activation_map}
\end{figure} 

% Problem that we are going to tackle is presented
%
Object detection (OD) is one of the most popular applications in the field of computer vision.
While a lot of efforts have been invested in developing efficient OD networks, state-of-the-art models still require huge computational resources. 
Therefore, practitioners are often forced to choose a suitable feature extractor (\aka{} backbone network), which has higher accuracy (\eg{} ResNet101) or faster speed (\eg{} MobileNetV2), depending on the amount of expendable resources on a target environment.

% Our approach to the problem is explained
%
In this paper, we propose an objectness-aware object detection method which processes only part of an input image where objects are likely to exist, namely \textit{foreground regions}.
This approach is motivated by the observation that most feature map computations in feature extraction network will be eventually discarded in anchor head network.
To mitigate this phenomenon, we build a light-weight objectness mask generation (OMG) network which creates a coarse-grained high-recall binary mask that distinguishes foreground regions from background.
This binary mask is then multiplied with a feature map before every convolution layer in an OD network.
Figure \ref{fig:activation_map} shows an example of (a) an input image, (b) a feature map generated by a modern OD model and (c) the proposed model.
While not a precise mask (\eg{} expanded boundaries, holes inside foreground regions), it covers most foreground regions.
As shown in (d), the proposed method detects all objects without background activation.
The proposed method can boost inference speed when combined with sparsity acceleration methods \cite{FacebookSparseConv_2018,NvidiaScnn_2017,sbnet_2018}.

% Briefly explain its effect
In experiments, we show that the proposed method effectively increases the number of zero values during forward pass with ReLU activation.
In case of MobileNetV2-SSDLite \cite{MobileNetV2_2018}, the original computational cost is 2.17 GMAC and its \textit{approximated GMAC (GMAC$^*$)} \footnote{GMAC$^*$ calculates the computational cost with only non-zero activation. See Equation \ref{eqn:mac} and \ref{eqn:mac_star} for detailed explanation.} is 1.61 on the MS-COCO \cite{MSCOCO_2014} validation dataset because of zero activation by ReLU.
When the proposed method is applied, GMAC$^*$ drops to 0.99 including both OMG and OD networks.
We found that this result is also consistent with heavy networks (VGG \cite{VGG_2015} and RetinaNet with ResNet101 \cite{FocalLoss}) and a different dataset (PASCAL VOC \cite{PascalVOCRetrospect_2015}).
In addition, models using the proposed method achieve comparable detection accuracy to their original counterparts.

The main contributions of this paper are three-fold:
\textbf{(i)} a novel object detection framework that reduces computational cost by sparsifying activation, not model parameters \cite{LotteryTicketHypothesis_2019,Han2015DeepCC,PruningWeightsAndConnections_2015,RethinkPruning_2019},
\textbf{(ii)} an end-to-end model architecture to incorporate the OMG network into a modern OD model and
\textbf{(iii)} experimental results show that training with only foreground region is not harmful.

%%%%%%%%%%%%%%%%%%%%%%%%%%%%%%%%%%%%%%%%%%%%%%%%%%%%%%%%%%%%%%%%
%%%%%%%%%%%%%%%%%%%%%%%%% Related Work %%%%%%%%%%%%%%%%%%%%%%%%%
%%%%%%%%%%%%%%%%%%%%%%%%%%%%%%%%%%%%%%%%%%%%%%%%%%%%%%%%%%%%%%%%

\section{Related Work}

% Related work on OD
Traditional object detection methods and OverFeat \cite{OverFeat_2014}, a deep learning approach, are based on the sliding window algorithm.
%% proposal-based two stage detector
To reduce the computational cost spent on searching the entire region with the sliding window manner, proposal-based two-stage R-CNN series have been proposed \cite{FastRCNN_2015,SPPNET_2015,FasterRCNN_2015}. 
It successfully reduces the number of candidates for the background region, however, it still requires lots of computing resources to generate proposals.
In the one-stage detectors such as SSD \cite{SingleShotDetector} and YOLO \cite{YOLO_2016}, the features of the convolutional backbone network are fed into subnetworks for object classification and bounding box regression. 
These one-stage detectors aim to be more efficient by directly classifying the pre-defined anchors and refining them using CNNs without the proposal generation step. 
However, it still requires to compute all of anchor candidates.

% Related work on OMG (semantic segmentation and saliency detection)
Semantic segmentation \cite{Cityscapes_2016,PascalVOCRetrospect_2015,MSCOCO_2014} is a task that classifies every pixel in an input image as one of pre-defined semantic classes.
Most previous work on this task \cite{DeepLab_2017,FullyConvNetForSS_2015,RecurrentConvNetForSS_2014,UNet_2015} utilize both high-level information on a large receptive field to recognize objects of complex shapes and low-level features to keep their spatial details.
For example, the seminal fully convolutional network model \cite{FullyConvNetForSS_2015} uses many convolutional and pooling layers to distill abstract information from an input image and bilinear up-sampling and skip-connections from lower layers to restore spatial details.
PSPNet \cite{PSPNet_2017} uses a pyramid pooling method to incorporate global context information.
Two or multi-branch systems \cite{FastSCNN_2019,ICNet_2018} are popular that process an input image of multiple scales in parallel and fuse their feature maps at the end.
However, the objective of these tasks is to generate a fine-grained segmentation map, which is different from our purpose that needs to build a coarse-grained high-recall binary mask as fast as possible.
Saliency detection \cite{BING_2014,SpectralResidual_2007,VisualSaliency_2010} is a similar task to semantic segmentation, except that salient areas are detected by pre-defined metrics not by training data.

% NN acceleration
%% network design, quantization, pruning, sparse conv deep 
Reducing the computational cost of neural networks can be tackled in various ways.
A straight-forward approach is to design efficient networks, manually \cite{PVANET_2016,MobileNetV2_2018} or automatically \cite{MobileNetV3_2019}.
From the opposite direction, we can train a heavy network and reduce model size by pruning insignificant parameters using network pruning techniques \cite{LotteryTicketHypothesis_2019,Han2015DeepCC,PruningWeightsAndConnections_2015,RethinkPruning_2019}.
These two approaches inevitably reduce network capacity and aggressive pruning suffers from accuracy drop.
Another approach is to use quantization method \cite{LowPrecisionDL_2017,BinaryConnect_2015,BinaryNet_2016}.
Computational costs rapidly decrease when low-bit precision data types are used for weights and activation.
Data sparsity can be exploited to increase the efficiency of neural networks.
For example, hardware/software-accelerated sparse convolution mechanism \cite{FacebookSparseConv_2018,NvidiaScnn_2017,sbnet_2018} greatly reduces the computational cost if weights or activation is very sparse.
Among them, SBNet \cite{sbnet_2018} is the most similar work to our approach, while they use a pipeline system with much heavier networks for 3D input data.
Several works \cite{SpatiallyAdaptiveComputation_2017,BranchyNet_2016,cascadedetection_2016} have proposed to dynamically scale computation through cascade or early termination.
Zhang \etal \cite{cascadedetection_2016} proposed cascade face detection which reduces face candidates by rejecting easy negatives in the early stage.
Michael \etal \cite{SpatiallyAdaptiveComputation_2017} proposed a method that adaptively determines after which layer to stop computation.
These approaches work effectively when the background is simple or not confused with the foreground. 
However, complex inputs require almost entire computation time, while the proposed method computes object region only.

%%%%%%%%% Approach %%%%%%%%%

\begin{figure*}[t]
\centering
\includegraphics[width=1.0\linewidth]{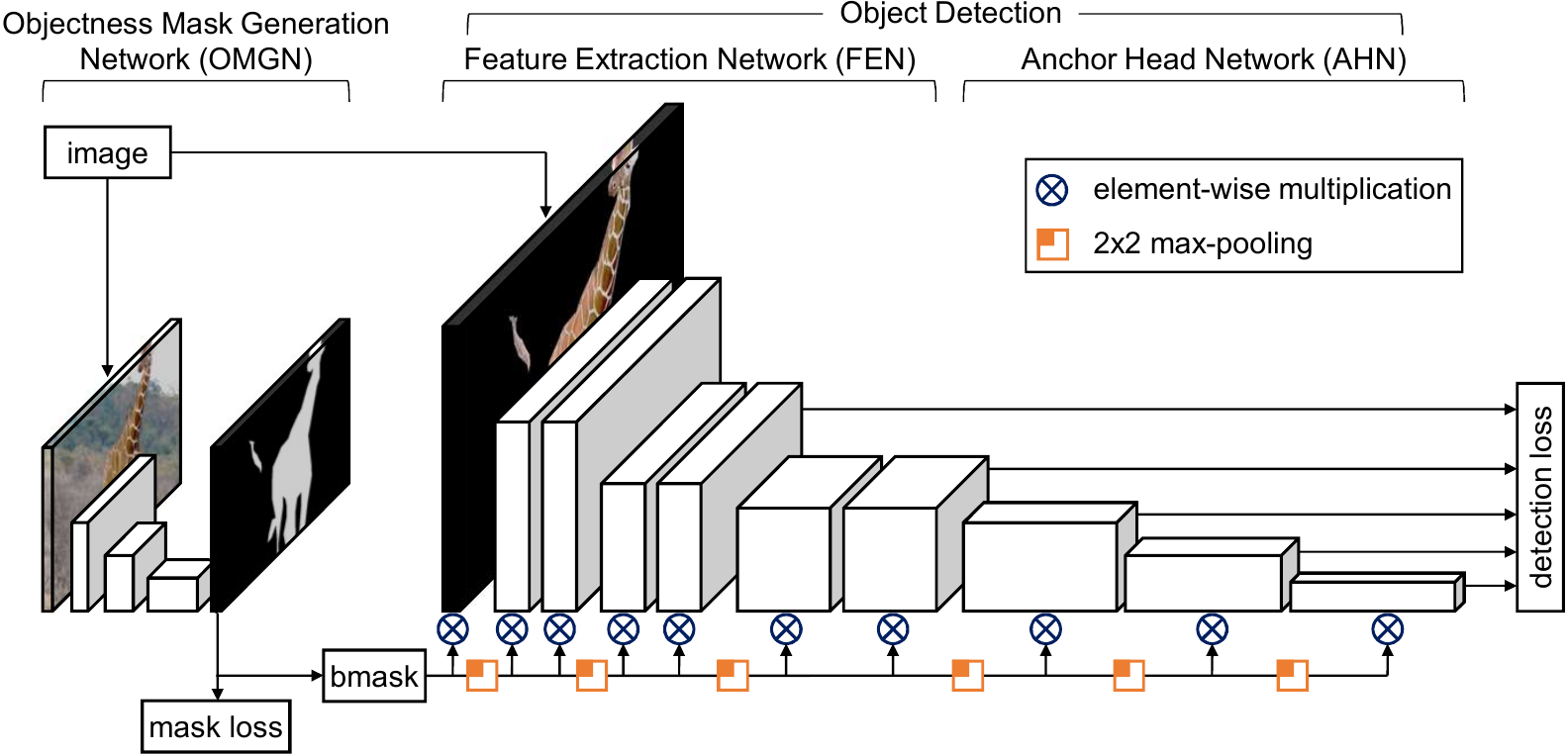}\vspace{-4mm}
\caption{An illustrative view of our method for SSD. We incorporate an OMG network in front of a modern object detector. An input image is first fed into the objectness mask generator and it produces a binary mask (\textit{bmask}). Then, the output feature map from each layer in the object detector (FEN+AHN) is multiplied with the mask to zero out the background region.}\vspace{-4mm}
\label{fig:e2e_network}
\end{figure*}

\section{Proposed Approach}

The objectness-aware object detection model consists of two modules: OMG network and OD network.
In Section \ref{sec:objectness_aware_od}, we illustrate the network architecture of the proposed model and its high level workflow.
Section \ref{sec:objectness_mask_generation} describes OMG network and Section \ref{sec:end_to_end_network} explains how to integrate these two networks into the end-to-end model in detail.

\subsection{Objectness-aware Object Detection}
\label{sec:objectness_aware_od}

% introduce base model od model
In this study, we use Single Shot MultiBox Detector (SSD) \cite{SingleShotDetector} as a baseline OD framework.
SSD consists of two main sub-modules: feature extraction network (FEN) and anchor head network (AHN).
The feature map located at each anchor position is fed into AHN to determine the object class as well as localize its bounding box.

% instance-aware od
The proposed objectness-aware OD model integrates an OMG network in front of the baseline OD network as shown in Figure \ref{fig:e2e_network}.
An input image and all input features to convolutional layers of the OD network are now masked with a binary objectness mask, which zeros out pixel values in the background area.
This masking operation is implemented as an element-wise multiplication layer denoted as $\otimes$ symbol in Figure \ref{fig:e2e_network}.
Also note that when the shape of a feature map changes due to the operations such as pooling and strided convolution, a max-pooling is applied to the mask so that its shape always matches that of the corresponding feature map.

The following section introduces our OMG method in detail.
We also evaluate other methods for objectness mask generation in Section \ref{exp:pipeline_approach}.

\subsection{Objectness Mask Generation}
\label{approach:omg}
\label{sec:objectness_mask_generation}
The OMG network takes an input image and generates a binary mask where background pixels have the value of zero and foreground pixels have the value of one.
This task can be thought of as a simplified version of semantic segmentation \cite{DeepLab_2017,FullyConvNetForSS_2015,RecurrentConvNetForSS_2014,FastSCNN_2019,UNet_2015,ICNet_2018,PSPNet_2017}.
The shape of an objectness mask can be an arbitrary shape, like a segmentation mask, or a box depending on the availability of annotation data.
We adopted Fast-SCNN \cite{FastSCNN_2019} as our OMG network because it is a fast and light-weight semantic segmentation model.
In addition, it is a fully convolutional network so that we can easily control its computational cost by changing the input image size.
In Figure \ref{fig:e2e_network}, the OMG network is shown in front of the OD network.

The purpose of OMG, however, is not identical to that of semantic segmentation.
For example, misclassification of the foreground area results in a more serious impact than the opposite, because the objects inside that region will be lost in the OD step.
In addition, even for a relatively large object, an OD model may experience difficulty in the regression of a bounding box position if the boundary areas (\eg{} arms or legs of a person) are lost.
Various methods are applied to emphasize the correct detection of foreground regions such as hard negative mining \cite{DeformablePartOD,FocalLoss,SingleShotDetector}, cost-sensitive learning \cite{ClassImbalanceLearning} and data pre-processing with dilation \cite{MorphologicalImageProcessing}.
In Section \ref{exp:omg}, we explain details of these methods and analyze their effects on the mask generation.

\subsection{Proposed Model}
\label{sec:end_to_end_network}

We incorporate OMG and OD networks into an end-to-end model to reduce the computational cost and improve the accuracy by directly optimizing two networks for the final goal.
However, it introduces a unique problem in the training process.
The OMG network generates a binary mask consisting of two discrete values by using $argmax$ function.
Because $argmax$ function is non-differentiable and the gradient is almost always zero, training the end-to-end model with the standard back-propagation is impossible.

There are two solutions we investigated to overcome this problem.
The first option is to use the soft-version of $argmax$ function,
\begin{equation}
    \label{eqn:softargmax}
    \text{soft-argmax}(x)=\sum_{i} \frac{e^{\beta x_{i}}}{\sum_{j} e^{\beta x_{j}}} i, 
\end{equation}
where $i$ is a class label and $\beta$ is a hyper-parameter that determines the shape of the resulting distribution.
We use $\beta$ = 5 in all experiments.
After training, this $soft\text{-}argmax$ function is replaced with $argmax$.
While this approach allows us to train the proposed model, the discrepancy between two activation functions yields an AP drop.

% surrogate gradient method
The second solution is to use the surrogate gradient (SG) method \cite{LowPrecisionDL_2017,BinaryNet_2016,UnderstandingSTE_2019}.
This method uses a proxy derivative function for the backward pass that approximates the direction of the original activation function's gradient.
We use
\begin{equation}
    \label{eqn:sg-argmax}
    \text{sg-argmax}(x)=
    \begin{cases}
        \text{argmax}(x) & \text{for forward} \\
        \text{soft-argmax}(x) & \text{for backward}
    \end{cases}
\end{equation}
that has differentiable and non-trivial gradient values suitable for back-propagation.
This technique has been widely used in training quantized neural networks where activation is a piece-wise constant function and shown successful results empirically.
In addition, a recent study \cite{UnderstandingSTE_2019} shows theoretical evidence that, when a proper proxy derivative is chosen, the training process converges near the original local minima.

%%%%%%%%% Experiments %%%%%%%%%

\section{Experiments} \label{experiments}
Section \ref{exp:preliminary}, \ref{exp:dataset_and_evaluation_metrics} and \ref{exp:implementation_details} explain terminologies, datasets and evaluation metrics, and implementation details used across experiments.
We conduct a preliminary experiment in Section \ref{exp:pipeline_approach} to investigate the best-known methods for training a good OMG model that plays a critical role in the proposed model.
We evaluate OMG models trained with various techniques and compare their performance to other objectness mask generation methods.
In Section \ref{exp:end_to_end_approach}, we integrate both OMG and OD networks into a single end-to-end model to explicitly guide the training process in the direction of improving the final object detection task performance.
Evaluation results with three types of networks and two datasets are reported.

\subsection{Terminologies}
\label{exp:preliminary}
A convolution layer consists of a triplet $\braket{X_i, X_o, W}$, where $X_i$ and $X_o$ correspond to the input and output activation tensors, respectively, and $W$ corresponds to the set of convolution kernels.
Each activation tensor is a three-dimensional tensor of size $w \times h \times c$, where $w$, $h$ and $c$ represent the tensor width, height, and depth, respectively.
For convenience, the dimensions of $X_i$ and $X_o$ are denoted by [$w_i \times h_i \times c_i$] and [$w_o \times h_o \times c_o$].
The set of convolution kernels $W$ is denoted by a four-dimensional tensor [$k \times k \times c_i \times c_o$], where $k$ is the kernel width and height.
The kernel width and height are not necessarily equal, but it is common practice in most conventional CNNs to take them.
Given the above notation, the number of multiply-accumulate (MAC) operations for a single convolution layer is defined as
\begin{equation}
    \label{eqn:mac}
    \text{MAC}=\frac{(w_i \times h_i \times c_i) \times c_o \times (k \times k)}{(s \times s)},
\end{equation}
where the $s$ is stride for convolution kernel and the padding is omitted for simplicity's sake. 
To calculate the MAC considering the zeros in activation tensor, we define approximated MAC as
\begin{equation}
    \label{eqn:mac_star}
    \text{MAC}^*=\frac{(w_i \times h_i \times c_i - z_i ) \times c_o \times (k \times k)}{(s \times s)},
\end{equation}
where the $z_i$ is the number of zeros in $X_i$ tensor.
Since MAC$^*$ approximates the MAC only considering the sparsity in activation tensor, the practical MAC can be reduced further if it also considers zero values in convolution kernels $W$, which can be maximized with network pruning \cite{Han2015DeepCC}.
We use the above notations throughout this paper.

\subsection{Dataset and Evaluation Metrics} 
\label{exp:dataset_and_evaluation_metrics}
Our experiments are performed on two benchmark datasets: MS-COCO \cite{MSCOCO_2014} and PASCAL VOC \cite{PascalVOCRetrospect_2015}.

% ms-coco
{\bf \noindent MS-COCO} contains bounding box and instance segmentation annotations for 80 categories and 115k images for training (\emph{train-2017}), 5k for validation (\emph{val-2017}) and 20k for testing without annotations (\emph{test-dev}).
We train models on \emph{train-2017}, report ablation studies on \emph{val-2017} and evaluate final performance on \emph{test-dev}.
The main evaluation metric is Average Precision (AP) that averages AP across IoU thresholds from 0.50 to 0.95 with an interval of 0.05.
Additionally, AP can be used to evaluate the performance under different object scales, including small objects (\emph{area} $<$ 32$^2$), medium objects (32$^2$ $<$ \emph{area} $<$ 96$^2$) and large objects (\emph{area} $>$ 96$^2$).

% pascal voc. voc2007:9,963 images / voc2012:11,530 images
{\bf \noindent PASCAL VOC} has bounding box annotations for 20 categories.
VOC2007 dataset contains 10k images for train, val and test and VOC2012 dataset has 11k images for train and val. We train models on VOC2007 and VOC2012 \emph{trainval} sets, and report results on VOC2007 \emph{test} set. Similar to the evaluation metrics used in the MS-COCO benchmark, we report the mean AP (mAP) averaged over all object categories with the IoU threshold 0.5.

\subsection{Implementation Details}
\label{exp:implementation_details}
For fair comparisons, all experiments are implemented based on publicly accessible softwares, MMDetection \cite{mmdetection2018} and Fast-SCNN \cite{FastScnnPytorch_Tramac}, using PyTorch.
The backbone networks used in our experiments are also publicly available.

In the pipeline approach, OMG models are trained for 160 epochs with one GPU and 64 batch size.
Initial learning rate is set to $0.02$ with the polynomial learning rate decay, which decreases the learning rate with $lr = lr_{initial} * (1 - epoch_{current} / epoch_{max})^{2}$.
Images of the training dataset are resized to fit their shorter edges to 640 pixels and randomly cropped into 320x320 patches.
OD models are trained with two GPUs (24 images per GPU) for 24 epochs with an initial learning rate of 0.02, decreased by a factor of 10 after 16 and 22 epochs.

To train end-to-end models, we use four GPUs (12 images per GPU) for 36 epochs with an initial learning rate of 0.02, decreased by a factor of 10 after 24 and 33 epochs.
The network backbones for OD are pre-trained on the ImageNet \cite{Russakovsky2015} and then fine-tuned on the detection dataset.
Because the OMG network is integrated into the OD network, its training scheme such as training epochs, learning rate, and data augmentation now follows that of OD models.
We increased the number of training epochs 1.5 times for the end-to-end models because the OMG network is not pre-trained at all and it affects OD network negatively at the beginning of the training.
All other hyper-parameters follow the settings in MMDetection \cite{mmdetection2018} and Fast-SCNN \cite{FastScnnPytorch_Tramac} if not mentioned in particular.

\subsection{Preliminary Experiments}
\label{exp:pipeline_approach}
%
% pipeline approach 
%
Objectness masks can be generated in many ways.
In this experiment, we build a pipeline system and runs a series of experiments using various mask generation methods including random mask generation, spectral residual saliency detection method \cite{SpectralResidual_2007} and the proposed OMG method.
We also use GT instance and bounding-box masks to investigate the upper-bound accuracy of the proposed approach.
In the following contents, we explain the proposed OMG method and object detection using various objectness masks in a pipeline manner.

\subsubsection{Objectness Mask Generation}
\label{exp:omg}

To make an OMG model better recognize foreground regions, we use the following three techniques.

\begin{enumerate}
\item {\bf \noindent Mask Dilation (MD)} expands the boundary of GT objectness masks using dilation \cite{MorphologicalImageProcessing} algorithm.
When an OMG model trained with dilated GT masks, it detects small and thin foreground areas better.
In Figure \ref{fig:dilation}, we can see that MD remarkably improves instance recall of predicted objectness masks at the cost of foreground ratio (\textit{FG-ratio}).

\begin{figure}
\centering
\includegraphics[width=0.62\linewidth]{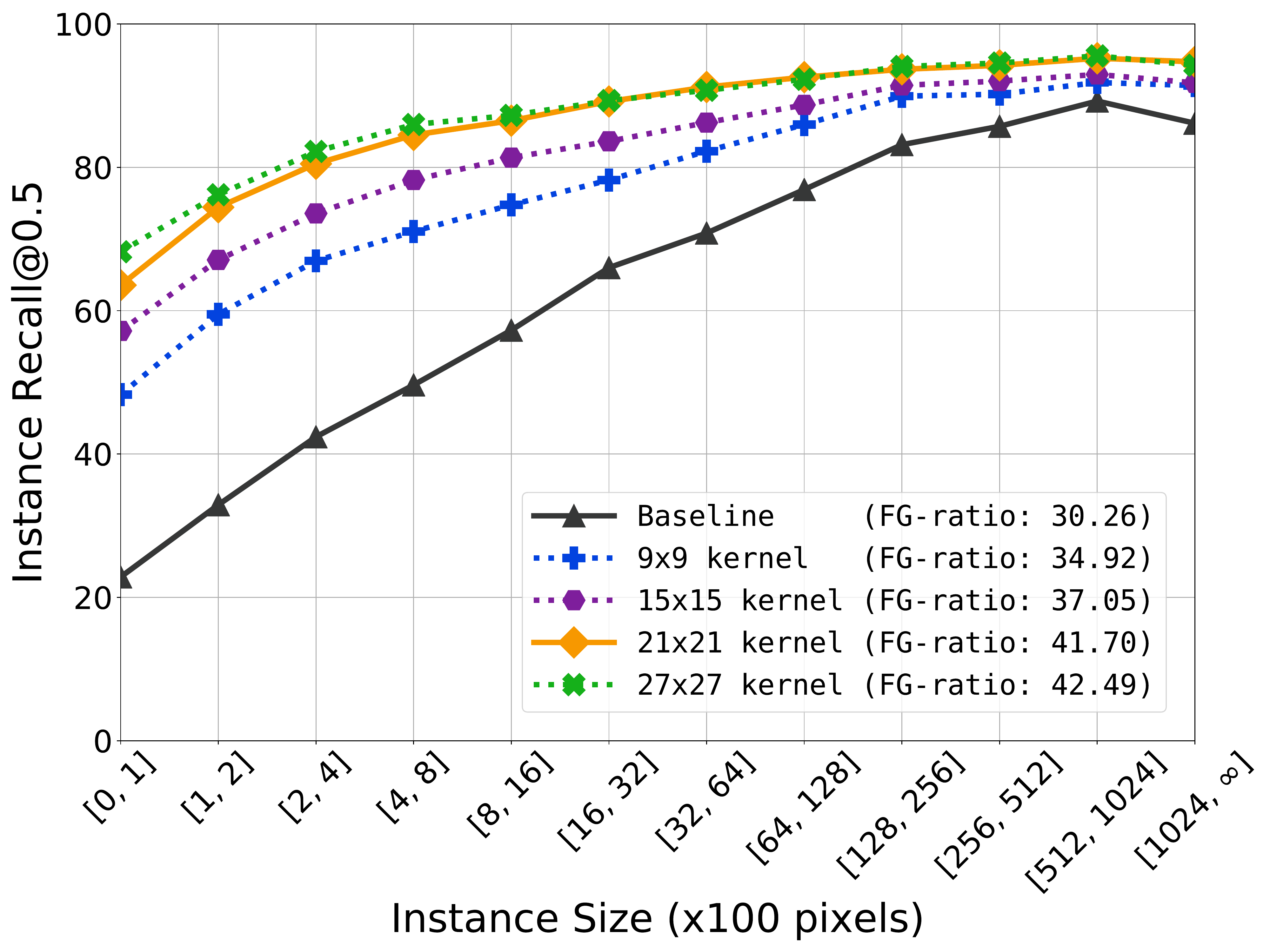}\vspace{-4mm}
\caption{The effect of mask dilation on MS-COCO \emph{val-2017} dataset. Note: X-axis shows the size of instances in specific ranges. Y-axis shows instance recall with the threshold 0.5 (if more than the half of the pixels of an instance is classified as foreground, that instance is considered as true positive).}\vspace{-3mm}
\label{fig:dilation}
\end{figure}

\item {\bf \noindent Online Hard-Negative Example Mining (OHNEM)} is a sampling method that selectively uses negative examples having the highest losses.
Because only a small fraction of the entire region is foreground in many images, the baseline model tends to predict background rather than foreground.
OHNEM mitigates this problem by using negative examples proportional to the number of positive examples.

\item {\bf \noindent Loss Re-Weighting (LW)} is a cost-sensitive learning method.
To emphasize the correct prediction of foreground areas, we give higher loss when they are misclassified.
We investigated how these methods influence the proposed OMG method.

\end{enumerate}

Table \ref{tab:omg_inst_recall} summarizes the results.
All of these methods improve instance recall at the cost of increasing FG-ratio.
We choose the combination of these three methods (\textit{ALL}) that maximizes instance recall while keeping the FG-ratio around 50\% and used for the E2E system.

\vspace{-5mm}
\begin{table}
\small
\centering
\begin{tabular}{c|c|c|c}
{\small Method} & {\small Param} & {\small Instance Recall@0.5} & {\small FG-ratio} \\ \hline \hline
Baseline & & 60.02 & 30.26 \\ \hline
\multirow{4}*{MD} & 9 & 75.54 & 34.92 \\
 & 15 & 80.67 & 37.05 \\
 & 21 & 85.73 & 41.70 \\
 & 27 & 86.68 & 42.49 \\ \hline
\multirow{3}*{OHNEM} & 1:3 & 66.54 & 32.62 \\
 & 1:1 & 73.97 & 38.67 \\
 & 2:1 & 85.98 & 55.02 \\ \hline
\multirow{2}*{LW} & 2:1 & 78.46 & 37.79 \\
 & 3:1 & 85.75 & 42.23 \\ \hline
ALL & 21, 1:3, 3:1 & 96.07 & 50.74 \\ \hline
\end{tabular}
\vspace{1mm}
\caption{The effect of MD, OHNEM, LW and their combination (ALL) on MS-COCO \emph{val-2017} dataset. \textit{Param} column indicates the kernel size, positive and negative sample ratio, and loss re-weighting values for positive and negative classes, respectively.}\vspace{-12mm}
\label{tab:omg_inst_recall}
\end{table}

\subsubsection{Object Detection}
\label{exp:pipeline_od}

In order to compare detection results using various objectness masks, we use MobileNetV2-SSDLite \cite{MobileNetV2_2018} as our baseline OD model, pre-trained on the ImageNet \cite{Russakovsky2015} dataset.
All models are fine-tuned on the MS-COCO \emph{train-2017} dataset with input image resolution of 512x512.
We perform evaluation on the \emph{val-2017} and report the results in Table \ref{tab:pipeline}.
All of these experiments except vanilla model use both images and pre-generated objectness masks.

\begin{table*}[t]
\small
\centering
\begin{tabular}{c|c|cc|c|cccccc}
mask & FG-ratio (\%) & zeros\% & zeros\% by mask & GMAC$^*$ & AP & AP\(_{50}\) & AP\(_{75}\) \\%& AP\(_{S}\) & AP\(_{M}\) & AP\(_{L}\)  \\ 
\hline\hline
vanilla     & 100 & 36.47 & {\footnotesize N/A}   & 1.61 & 22.6 & 40.3 & 22.7 \\%& 6.7 & 23.7 & 36.5 \\ 
\hline
GT mask     & 30  & 81.66 & 66.64 & 0.54 & 35.8 & 51.3 & 38.5 \\%& 15.9 & 38.8 & 48.5 \\
GT box mask & 43  & 72.56 & 54.03 & 0.74 & 37.6 & 51.3 & 40.0\\% & 15.9 & 40.8 & 53.2  \\ 
\hline
SR mask     & 50  & 66.50 & 43.45 & 0.97 & 16.8 & 32.2 & 15.7 \\%& 4.5 & 18.0 & 26.3  \\
random mask & 30  & 47.78 &  9.90 & 1.51 & 17.5 & 32.0 & 17.1 \\%& 3.2 & 17.0 & 30.3  \\
random mask & 50  & 42.92 &  3.80 & 1.54 & 18.6 & 34.0 & 18.2 \\%& 3.8 & 18.8 & 31.5  \\ 
\hline
OMG mask    & 51  & 69.02 & 46.39 & 0.88 & 20.4 & 36.5 & 20.4 \\%& 4.7 & 20.6 & 34.6  \\ 
\hline \end{tabular}
\vspace{1mm}
\caption{Preliminary experiments: comparison of different static objectness mask for objectness-aware object detection on MS-COCO \emph{val-2017}.}\vspace{-7mm}
\label{tab:pipeline}
\end{table*}

\begin{enumerate}
\item
{\bf \noindent Vanilla.}
The vanilla model has 36.47\% zero values in the entire feature maps because of the ReLU activation \cite{ReLU_2010}.
Similar to the reported performance in the original paper \cite{MobileNetV2_2018}, it achieves 22.6 AP as shown in Table \ref{tab:pipeline}.

\item
{\bf \noindent Instance GT Mask.}
To understand the effect of using only foreground region for object detection and its upper-bound performance, we trained and evaluated a model using the GT instance segmentation annotation as objectness masks.
Without bells and whistles, it achieves 35.8 AP which is 13.2 points higher than the vanilla model.
It implies that background information is not necessarily essential and high-quality masks can help to achieve better accuracy for object detection.
In terms of computational cost, it requires only 33.5\% (0.54 GMAC$^*$) of the vanilla model (1.61 GMAC$^*$).

\item
{\bf \noindent Box GT Mask.}
Since building an instance segmentation dataset is time consuming and costly work, not all datasets have instance segmentation annotation.
Assuming that a given dataset has only bounding-box annotation, we trained a model to examine the upper-bound performance of the proposed method.
It achieves the highest AP (37.6) among all models, even higher than the instance GT mask model.
We speculate that the model performs better because the shape of objectness masks are identical to that of target bounding boxes.
However, its average FG-ratio is 13\% higher than that of the instance GT mask model.

\item
{\bf \noindent SR Mask.}
Spectral residual method \cite{SpectralResidual_2007} is one of saliency detection algorithms.
It identifies specific areas of an image where unusual frequency information resides based on the natural image statistics.
This method doesn't need a training dataset and runs very fast.
We trained a model using objectness masks generated with this method.
To limit the average FG-ratio under 50\%, the threshold 0.08 is used.
This model achieves the lowest AP among all models, even worse than the models trained with random masks.
It results from the fact that salient areas don't necessarily correspond to the regions where target objects exist.

\item
{\bf \noindent Random Mask.}
One of the simplest approaches to reduce MAC$^*$ is to use random objectness masks.
We trained two models with random masks generated by using a salt-and-pepper noise pattern with targeted FG-ratios, 30\%, and 50\%.
Compared to the baseline model, these models yield noticeable accuracy drop, 5.1 and 4.0 points, respectively.
However, the biggest problem is the GMAC$^*$ values.
While the average FG-ratios are 30\% and 50\%, their GMAC$^*$ values aren't much different from that of the vanilla model because random background pixels mostly disappear when the first max-pooling is applied to reduce mask shape.

\item
{\bf \noindent OMG Mask.}
We trained a model with the objectness masks generated by the proposed OMG method explained in \ref{approach:omg}.
Unlike the training process, raw input images are used without resizing for mask generation in the same manner as \textit{SR Mask} and \textit{Random Mask} models.
This is different from our end-to-end model that uses resized input images.
Although it yields 2.2 lower AP compared with the vanilla model, its GMAC$^*$ is almost half of it.
In the following section, we present the experiments using the end-to-end model that mitigates this accuracy gap. \vspace{-3mm}

\end{enumerate}

\begin{figure}[t]
\centering
\includegraphics[width=0.62\linewidth]{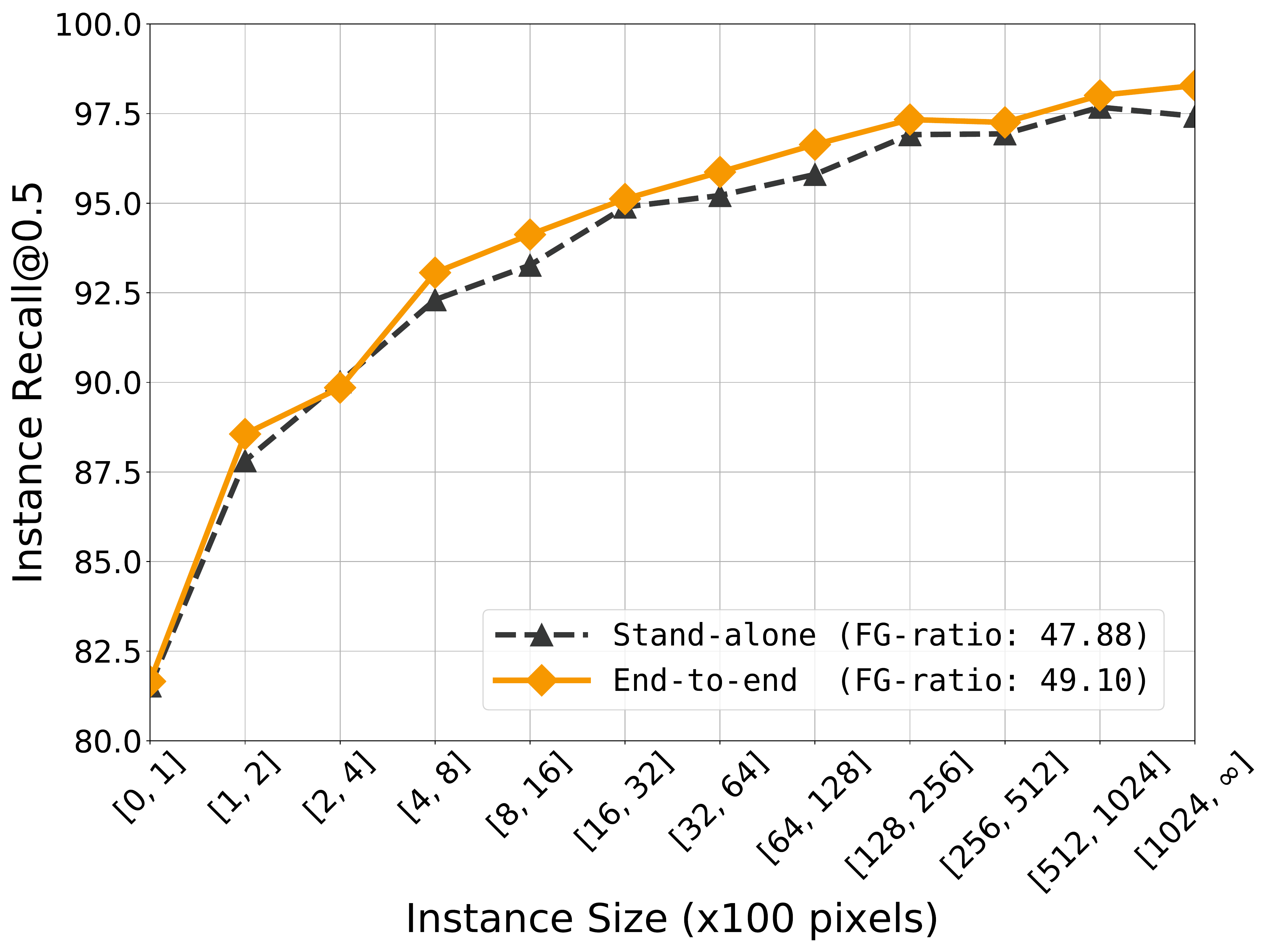}\vspace{-4mm}
\caption{Comparison of instance recall of OMG masks generated from the stand-alone OMG model and the end-to-end model. These models use the same training and inference schemes such as input image size and data augmentation methods.}\vspace{-3mm}
\label{fig:omg-only-vs-e2e-masks}
\end{figure}

\subsection{Proposed Approach}
\label{exp:end_to_end_approach}

In the case of MobileNetV2, we use an input image resolution of 256$^2$ for OMG and 512$^2$ for OD to reduce the computational cost for mask generation.
The generated mask is re-scaled to 512$^2$ by bilinear up-sampling before $argmax$ output. 
The computational cost of OMG network is only 9.2\% of MobileNetV2-SSDLite.

To compare the quality of objectness masks between the pipeline and the proposed end-to-end approaches, we evaluate two models with the same configuration except for the connection between OMG and OD networks.
In Figure \ref{fig:omg-only-vs-e2e-masks}, the masks generated from the end-to-end model exhibit higher instance recall than those generated by the pipeline model for all instance sizes.
This experiment result shows that additional losses back-propagated from the OD network to the OMG network contribute to the high-quality mask generation.

Our end-to-end model also achieves 1.2 points higher AP than the pipeline model as shown in Table \ref{tab:omg_input_size}.
It implies that improved mask generation by OD losses also helps to enhance object detection performance.
In addition, we tried a larger input image size (512$^2$) for the OMG network.
It improves 0.2 points in AP, but increases GMAC$^*_{OMG}$ four times from 0.14 to 0.56.

\begin{figure}
\centering
\includegraphics[width=0.7\linewidth]{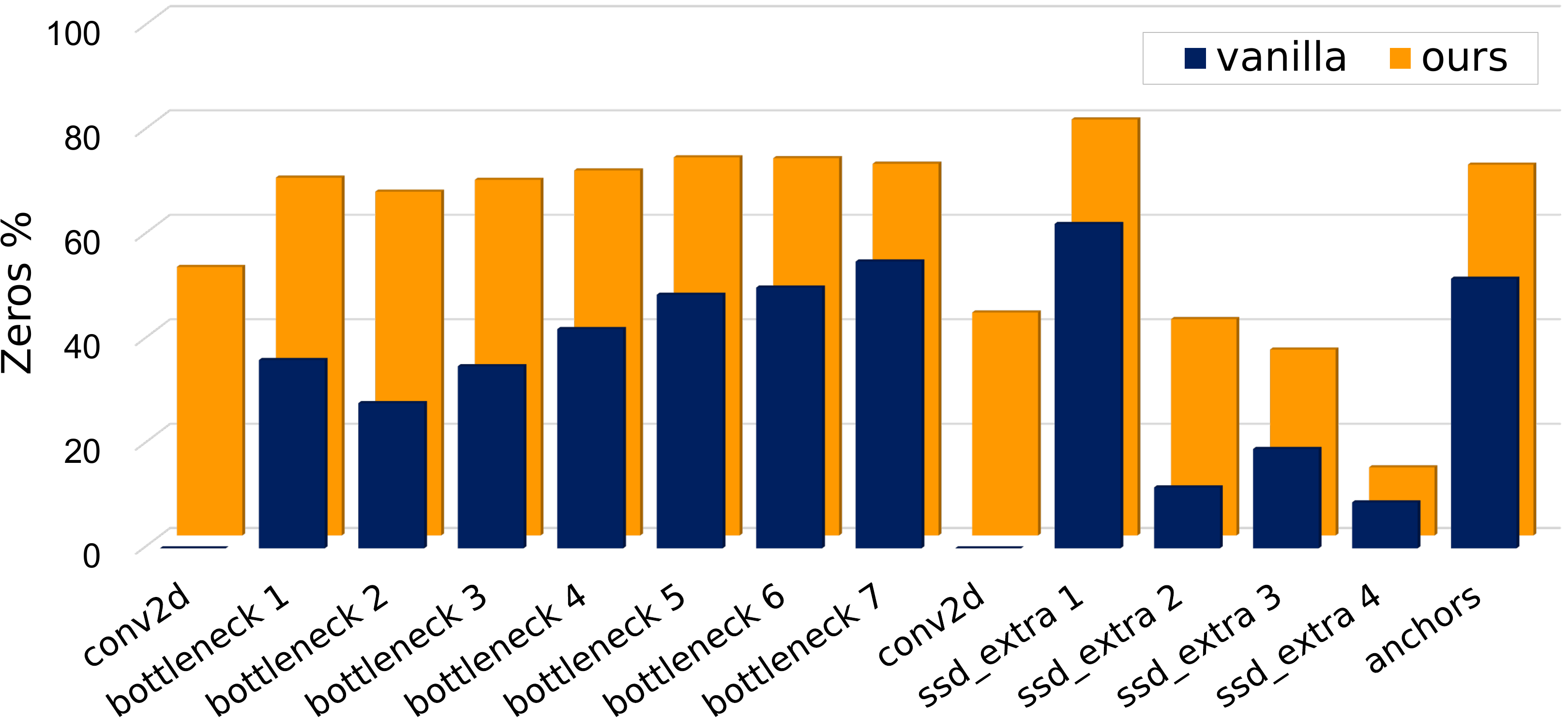}\vspace{-2mm}
\caption{The proportion of zero values at each feature map blocks in the object detection network (MobileNetV2-SSDLite); our method largely increases zero values for the entire feature maps on MS-COCO \emph{val-2017}.}\vspace{-2mm}
\label{fig:zero_values}
\end{figure}

Figure \ref{fig:zero_values} shows the average ratio of zero values at each feature map blocks for MS-COCO validation set compare to the vanilla model.
Although ReLU activation already made activation very sparse (36.47\% of feature maps are zero in the vanilla model), the proposed method pushes this number to 68.10\%.
This leads to the reduction of computational cost by 0.61 GMAC$^*$, 37.89\% decrease.

\begin{table}[t]
\small
\setlength{\tabcolsep}{0.5em}
\centering
\begin{tabular}{c|c|c|cc|ccc}
\multirow{2}*{\footnotesize OMG input}%\thead{OMG\\input}} 
& \multirow{2}*{\small E2E} & \multicolumn{3}{c|}{\small GMAC$^*$} & \multicolumn{3}{c}{\small COCO} \\
\cline{3-8} & & {\small sum} & {\small OMG} & {\small OD} & {\small AP} & {\small AP\(_{50}\)} & {\small AP\(_{75}\)} \\ \hline \hline
256$^2$ & {\small \ding{56}} & 0.95 & 0.14 & 0.81 & 20.2 & 36.1 & 20.4 \\ \hline
256$^2$ & {\small \ding{52}} & 1.01 & 0.14 & 0.87 & 21.4 & 37.6 & 21.5 \\ \hline
512$^2$ & {\small \ding{52}} & 1.39 & 0.56 & 0.83 & 21.6 & 37.8 & 21.7 \\ \hline
\end{tabular}
\vspace{1mm}
\caption{Results of different combinations of the OMG network.}\vspace{-5mm}
\label{tab:omg_input_size}
\end{table}

\begin{table*}[t]
\centering
\small
\begin{tabular}{c|c|c|c|c|c|c|cccccc}
& {\footnotesize width multiplier} & {\small image} & {\small mask GT} & {\small SG} & {\small GMAC} & {\small GMAC$^*$} & {\small AP} & {\small AP\(_{50}\)} & {\small AP\(_{75}\)} \\ \hline\hline
%& {\small AP\(_{S}\)} & {\small AP\(_{M}\)} & {\small AP\(_{L}\)}  \\ \hline\hline

\multirow{5}*{\small vanilla} 
& 1.0 & 512$^2$ & \multirow{5}*{\footnotesize N/A} & \multirow{5}*{\footnotesize N/A} & \textbf{2.17} & \textbf{1.61} & \textbf{22.6} & \textbf{40.3} & \textbf{22.7} \\%& 5.0 & 20.0 & 32.6 \\
& 0.75 & 512$^2$ &  &  & 1.61 & 1.21 & 20.0 & 36.5 & 19.4 \\%& 5.0 & 20.0 & 32.6 \\
& 0.5  & 512$^2$ &  &  & 0.95 & 0.69 & 15.9 & 30.0 & 15.1 \\%& 3.1 & 15.4 & 27.5 \\
& 1.0  & 416$^2$ &  &  & 1.44 & 1.07 & 20.1 & 37.0 & 19.4 \\%& 4.8 & 19.8 & 33.6 \\
& 1.0  & 352$^2$ &  &  & 1.03 & 0.77 & 17.8 & 33.8 & 16.9 \\%& 4.0 & 17.6 & 29.0 \\
\hline
\multirow{3}*{ours} & \multirow{3}*{1.0} & \multirow{3}*{512$^2$}
    & {\small instance} & {\small \ding{56}} & \multirow{3}*{2.37} & 1.26 & 21.2 & 37.9 & 21.3 \\%& 5.0 & 21.8 & 35.5 \\
& & & {\small instance} & {\small \ding{52}} & & 1.01 & 21.4 & 37.6 & 21.5 \\%& 5.6 & 21.8 & 35.9 \\

& & & {\small box     } & {\small \ding{52}} & & 1.17 & 22.0 & 38.3 & 22.2 \\ \hline%& 5.6 & \textbf{22.7} & 36.4 \\ \hline
ours$^+$ & 1.0 & 512$^2$ & {\small instance} & {\small \ding{52}} & \textbf{2.37} & \textbf{0.99} & \textbf{22.7} & \textbf{39.4} & \textbf{23.1} \\ \hline%& \textbf{6.4} & 23.1 & \textbf{38.2} \\ \hline

\end{tabular}
\vspace{1mm}
\caption{Comparison of the proposed method to the other MAC-reduction approaches on MS-COCO \emph{val-2017}}\vspace{-6mm}
\label{tab:mv2_e2e_results}
\end{table*}

\vspace{-4mm}

In Table \ref{tab:mv2_e2e_results}, we compare the proposed method with other MAC-reduction approaches using low resolution input images and small capacity networks (with \textit{width multiplier} option of MobileNetV2).
As shown in Table \ref{tab:mv2_e2e_results}, our models using the proposed method always outperform the vanilla models, even if some of them have larger MAC$^*$ than ours.
In addition, the use of SG method (Eq. \ref{eqn:sg-argmax}) improves both AP and GMAC$^*$ than \emph{soft-argmax} method (Eq. \ref{eqn:softargmax}).

In addition, we trained the model three times more epochs to see if it has converged completely.
The trained model, shown as $ours^+$ in Table \ref{tab:mv2_e2e_results}, achieved 22.7 points AP (+1.5) and 0.99 GMAC$^*$ (-0.02).
This result shows that training the end-to-end model requires additional hyper-parameter tuning.

\begin{table}[t]
\small
\centering
\setlength{\tabcolsep}{0.46em}
\begin{tabular}{c|c|c|c|ccc}
{\small network} & {\small method} & {\small GMAC} & {\small GMAC$^{*}$} & {\small AP} & {\small AP\(_{50}\)} & {\small AP\(_{75}\)} \\ \hline\hline
% 512x512
\multirow{3}*{VGG} & vanilla & 86.70 & 37.38 & 29.5 & 48.8 & 30.9 \\
\cline{2-7}        & GT mask & 86.70 & 11.85 & 44.5 & 61.5 & 48.5 \\ 
\cline{2-7}        & ours    & 87.53 & 23.91 & 27.5 & 46.2 & 28.5 \\ \hline
% 1216x800
\multirow{3}*{\thead{RetinaNet\\w/ R101}} 
                  & vanilla & 135.02 & 53.56 & 37.6 & 57.6 & 40.1 \\
\cline{2-7}       & GT mask & 135.02 & 18.24 & 51.3 & 67.1 & 55.1 \\ 
\cline{2-7}       & ours    & 137.89 & 37.06 & 36.5 & 56.3 & 39.0 \\ \hline
\end{tabular}
\vspace{2mm}
\caption{Evaluation results of the end-to-end model with heavy backbones on MS-COCO \emph{val-2017}. The input image resolution of 512x512 and 1216x800 is used for VGG and RetinaNet, respectively.}
\label{tab:heavy_backbone}
\end{table}

\begin{table}[t]
\small
\centering
\setlength{\tabcolsep}{0.4em}
\vspace{-2mm}
\begin{tabular}{c|c|c|c|ccc} 
{\small network} & {\small method} & {\small GMAC} & {\small GMAC$^*$} & {\small AP} & {\small AP\(_{50}\)} & {\small AP\(_{75}\)} \\ \hline \hline
\multirow{2}*{\small MobileNet} & {\small vanilla} & 2.17 & 1.61 & 22.8 & 40.6 & 22.8 \\
\cline{2-7}  & {\small ours} & 2.37 & 1.00 & 21.5 & 37.9 & 21.4 \\ \hline
\multirow{2}*{\small VGG} & {\small vanilla} & 86.70 & 37.35 & 29.6 & 49.0 & 31.2 \\
\cline{2-7}  & {\small ours} & 87.53 & 23.65 & 27.6 & 46.2 & 28.9 \\ \hline
\multirow{2}*{\small RetinaNet} & {\small vanilla} & 135.02 & 53.52 & 38.0 & 58.5 & 40.9 \\
\cline{2-7}  & {\small ours} & 137.89 & 36.70 & 36.7 & 56.9 & 39.4 \\ \hline
\end{tabular}
\vspace{1mm}
\caption{Evaluation results on MS-COCO \emph{test-dev}}\vspace{-7mm}
\label{tab:test_dev}
\end{table}

\begin{table}[t]
\small
\centering
\vspace{-2mm}
\begin{tabular}{c|c|cc|c}
{\small method} & {\small FG-ratio (\%)} & {\small GMAC} & {\small GMAC$^*$} & {\small mAP} \\ \hline \hline
vanilla     & 100 & 1.82 & 1.39 & 74.0 \\ \hline
GT box mask & 44  & 1.82 & 0.67 & 80.1 \\ \hline
ours        & 64  & 2.03 & 1.03 & 71.4 \\ \hline
\end{tabular}
\vspace{1mm}
\caption{Evaluation results on VOC2007 \emph{test} }\vspace{-3mm}
\label{tab:voc}
\end{table}

\begin{figure*}
\centering
\def\arraystretch{0.8}
\setlength{\tabcolsep}{0.07em}
\begin{tabular}{ccccccccccc}

\multirow{1}{*}[13.5ex] {\rotatebox{90}{\scriptsize result (vanilla)}} &
\multicolumn{2}{c}{
\begin{overpic}[width=0.19\textwidth]{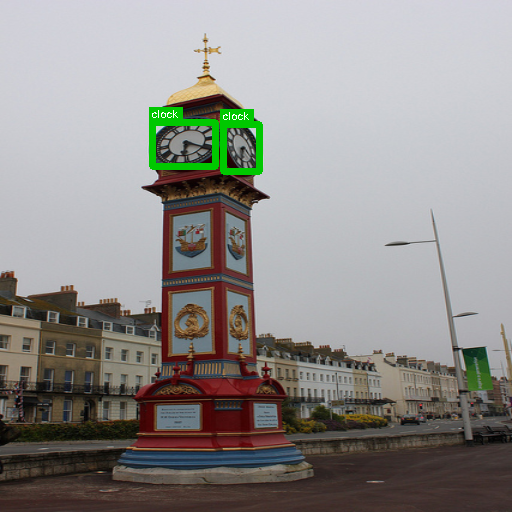}
\put (48, 3) {\small \color{white} \textbf{1.62} G$^*$}
\end{overpic}} & 
\multicolumn{2}{c}{
\begin{overpic}[width=0.19\linewidth]{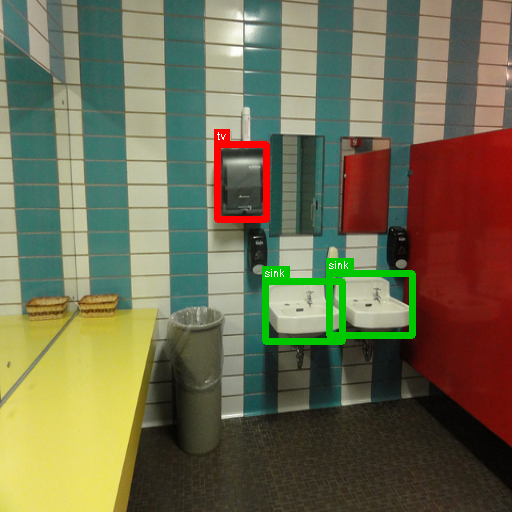} 
\put (48, 3) {\small \color{white} \textbf{1.61} G$^*$}
\end{overpic}} &
\multicolumn{2}{c}{
\begin{overpic}[width=0.19\linewidth]{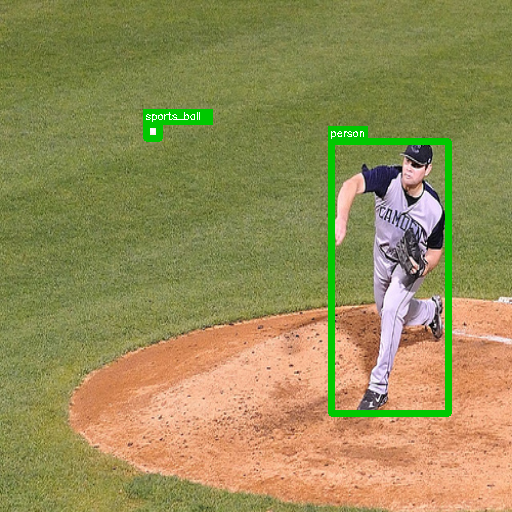} 
\put (48, 3) {\small \color{white} \textbf{1.62} G$^*$}
\end{overpic}} &
\multicolumn{2}{c}{
\begin{overpic}[width=0.19\linewidth]{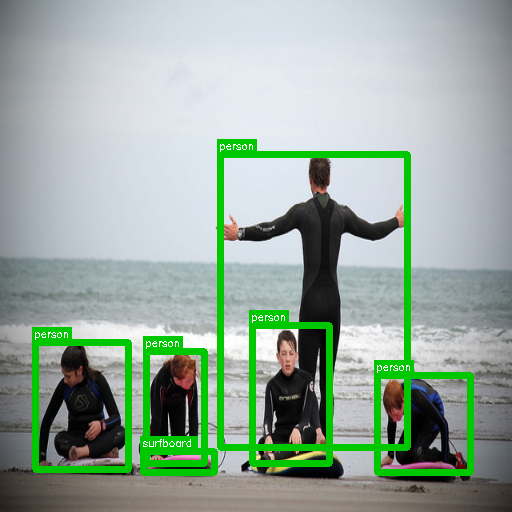} 
\put (48, 3) {\small \color{white} \textbf{1.62} G$^*$}
\end{overpic}} &
\multicolumn{2}{c}{
\begin{overpic}[width=0.19\linewidth]{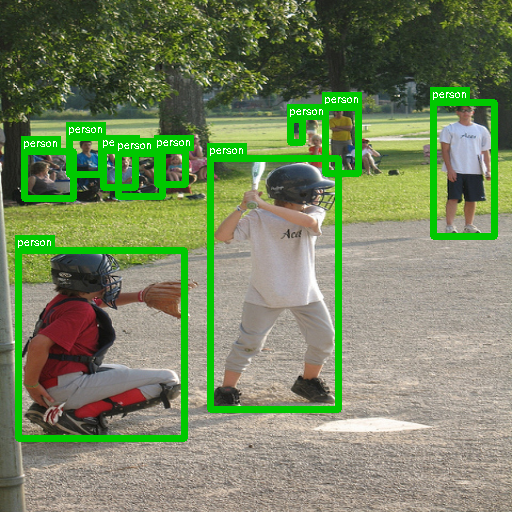} 
\put (48, 3) {\small \color{white} \textbf{1.61} G$^*$}
\end{overpic}}\\

\multirow{7}*[0ex] {\rotatebox{90}{\scriptsize feature map}} &
\multirow{5}*{\includegraphics[width=0.093\linewidth]{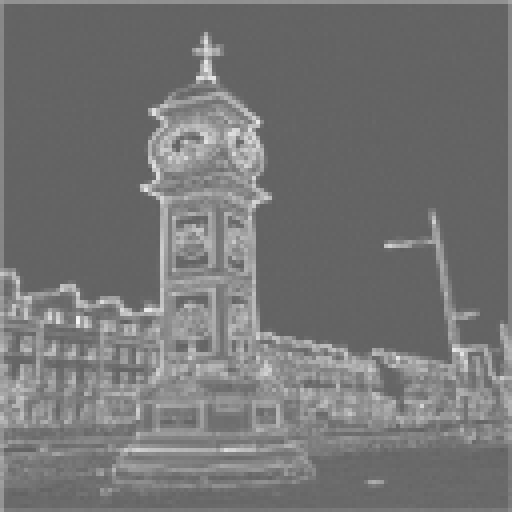}} &
& \multirow{5}*{\includegraphics[width=0.093\linewidth]{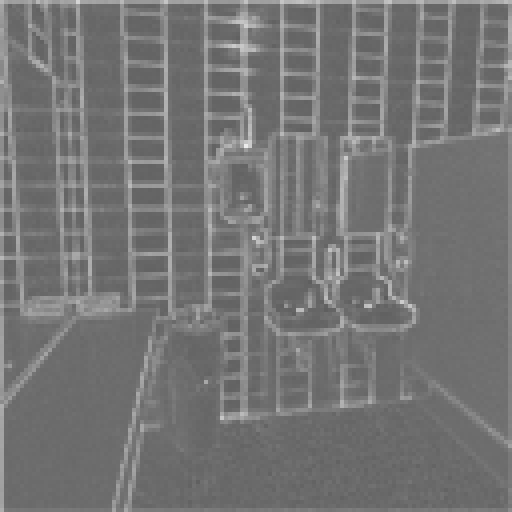}} &
& \multirow{5}*{\includegraphics[width=0.093\linewidth]{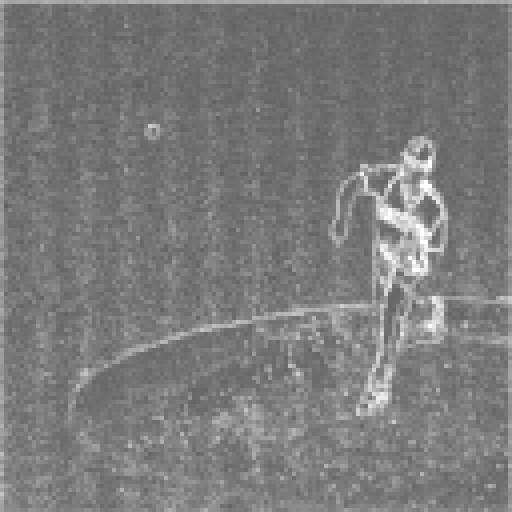}} &
& \multirow{5}*{\includegraphics[width=0.093\linewidth]{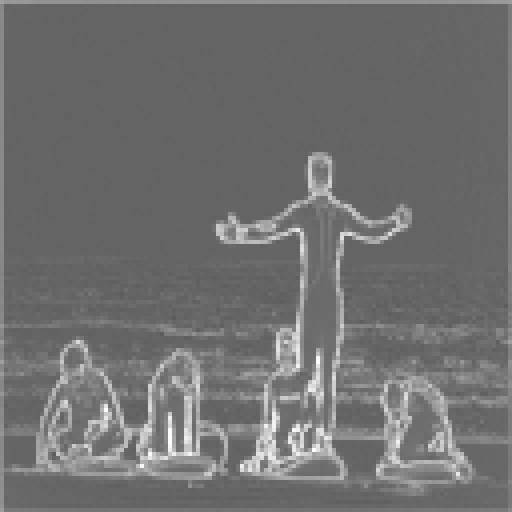}} &
& \multirow{5}*{\includegraphics[width=0.093\linewidth]{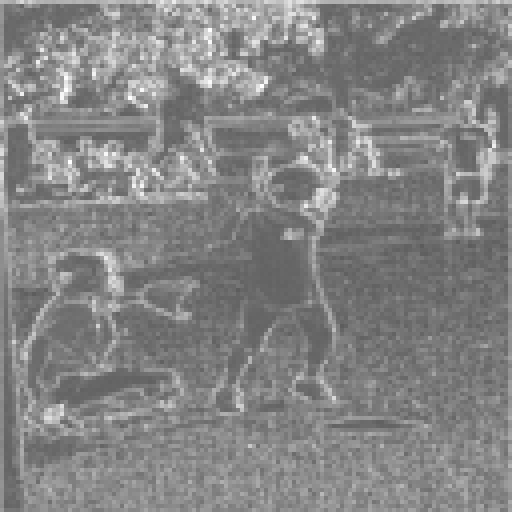}} & \multirow{2}*{{\tiny $\leftarrow{}$}{\scriptsize vanilla}} \\
& & & & & & & & & & \\
& & \multirow{5}*{\includegraphics[width=0.093\linewidth]{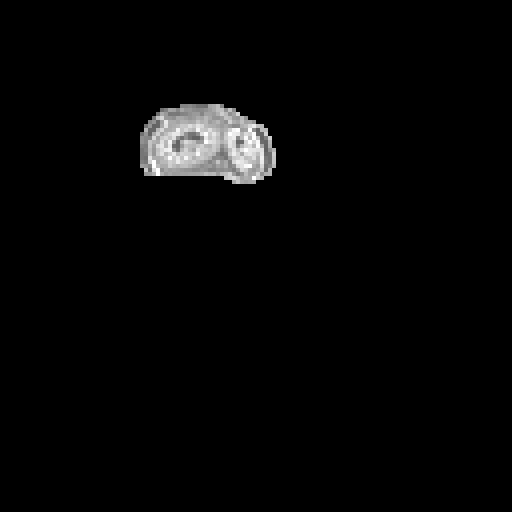}} &
& \multirow{5}*{\includegraphics[width=0.093\linewidth]{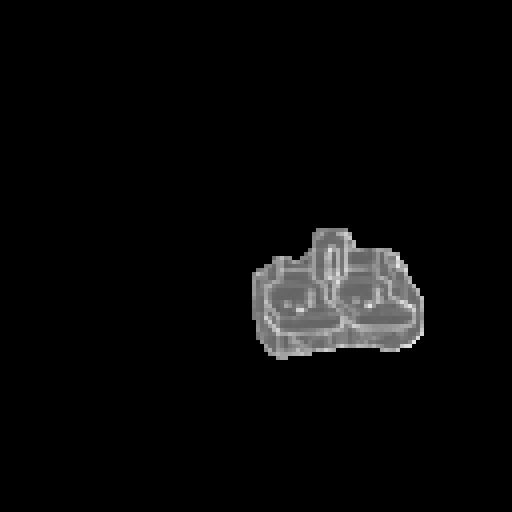}} &
& \multirow{5}*{\includegraphics[width=0.093\linewidth]{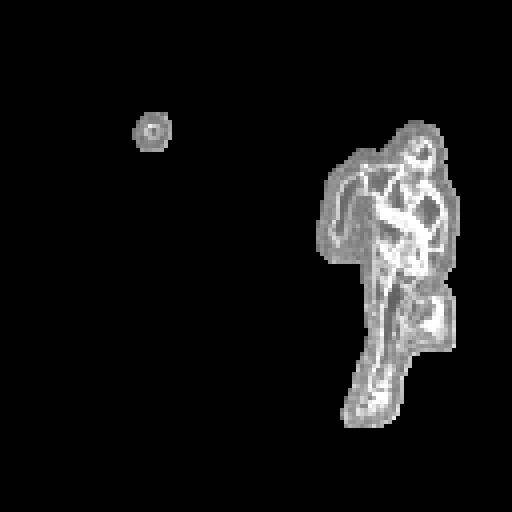}} &
& \multirow{5}*{\includegraphics[width=0.093\linewidth]{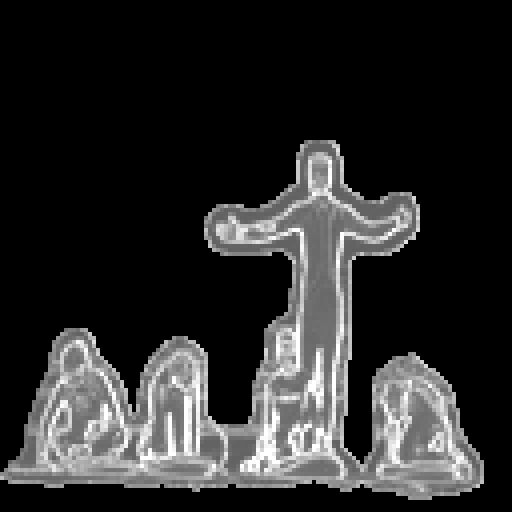}} &
& \multirow{5}*{\includegraphics[width=0.093\linewidth]{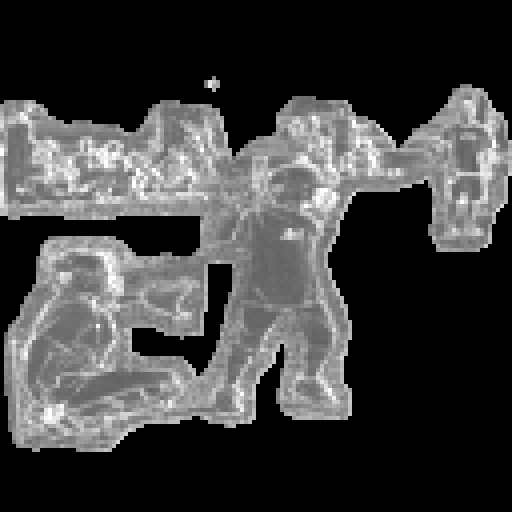}} \\
& & & & & & & & & & \\
& & & & & & & & & & \\
& \multirow{2}*{{\scriptsize ours} {\tiny $\rightarrow{}$}} & & & & & & & & & \\
& & & & & & & & & & \\

\multirow{1}{*}[12.5ex] {\rotatebox{90}{\scriptsize result (ours)}} &
\multicolumn{2}{c}{
\begin{overpic}[width=0.19\linewidth]{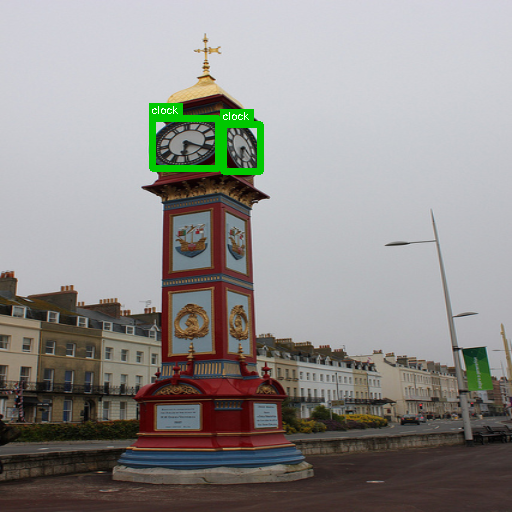} 
\put (48, 3) {\small \color{white} \textbf{0.21} G$^*$}
\put (68, 88) {\small \color{white} \textbf{7.7$\times$}}
\end{overpic}} &
\multicolumn{2}{c}{
\begin{overpic}[width=0.19\linewidth]{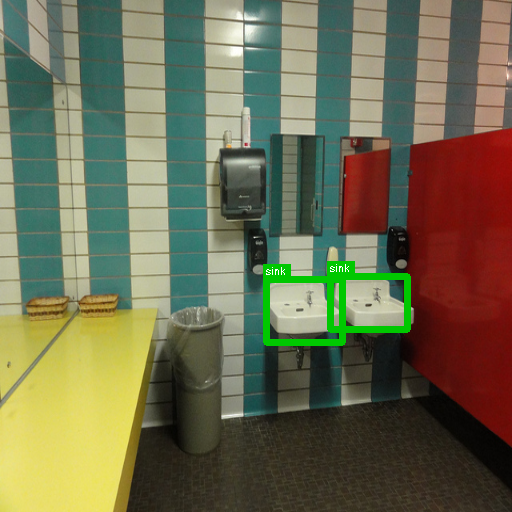} 
\put (48, 3) {\small \color{white} \textbf{0.27} G$^*$}
\put (68, 88) {\small \color{white} \textbf{6.0$\times$}}
\end{overpic}} &
\multicolumn{2}{c}{
\begin{overpic}[width=0.19\linewidth]{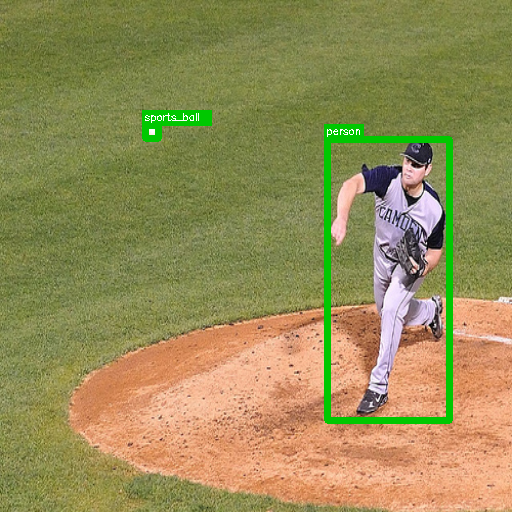} 
\put (48, 3) {\small \color{white} \textbf{0.38} G$^*$}
\put (68, 88) {\small \color{white} \textbf{4.3$\times$}}
\end{overpic}} &
\multicolumn{2}{c}{
\begin{overpic}[width=0.19\linewidth]{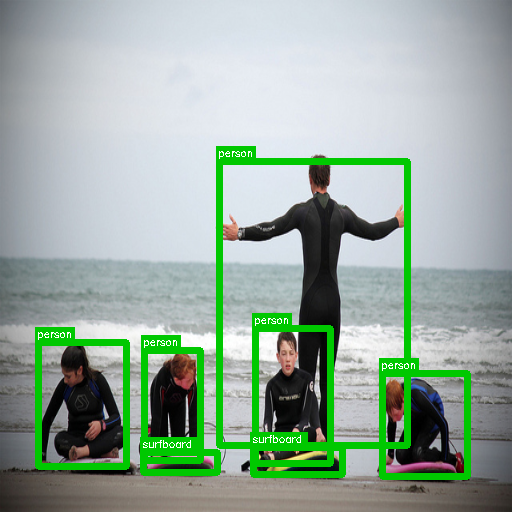} 
\put (48, 3) {\small \color{white} \textbf{0.72} G$^*$}
\put (68, 88) {\small \color{white} \textbf{2.3$\times$}}
\end{overpic}} &
\multicolumn{2}{c}{
\begin{overpic}[width=0.19\linewidth]{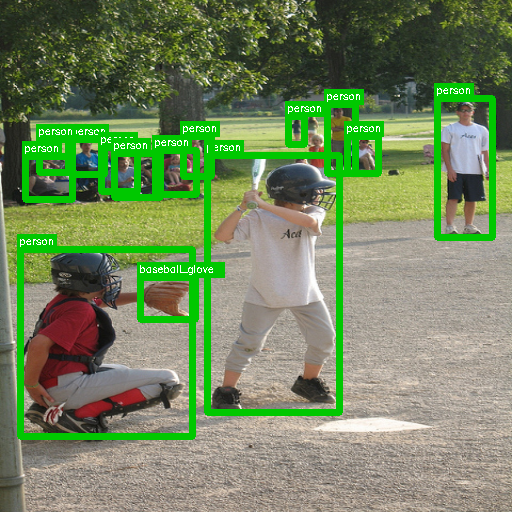} 
\put (48, 3) {\small \color{white} \textbf{0.95} G$^*$}
\put (68, 88) {\small \color{white} \textbf{1.7$\times$}}
\end{overpic}}

\end{tabular}
\vspace{-3mm}
\caption{Qualitative comparison between the vanilla model and the proposed model(end-to-end). {\color{green} Green} box represents true positive, whereas {\color{red} red} box indicates a false positive. (G$^*$ stands for GMAC$^*$ and it includes both OMG and OD network's computations)}\vspace{0mm}
\label{fig:quailitative}
\end{figure*}

Figure \ref{fig:quailitative} shows sample images with their feature maps and detection results for the vanilla model and the proposed model.
To visualize a feature map, we average absolute feature values across channels at the end of bottleneck block 2 which has 128$^2$ size.
As shown in the second row, our method generates a feature map where only a small portion is the foreground region.
As a result, MAC$^*$ value for each image is very small as shown in the right bottom area of that image.
These results tell us that the proposed model is going to operate with very low MAC$^*$ in many practical applications such as autonomous driving or surveillance camera.
In addition, the objectness mask often filters out false-positives of the vanilla model, as shown in the second column image.

To evaluate the generality of the proposed method, we have experimented with heavy backbone networks.
Table \ref{tab:heavy_backbone} shows the MAC$^*$ reduction comparing to their vanilla models.
Our approach only uses 64\% and 69\% of the vanilla GMAC$^*$ for the VGG and RetinaNet models respectively, while experiencing a little drop of accuracy.
In addition, we conjecture that the accuracy drop can be mitigated by training more epochs as the experiment of MobileNetV2 ($ours^+$).

Lastly, Table \ref{tab:test_dev} reports the evaluation result on MS-COCO \emph{test-dev}.
The result of the three models is almost consistent with the experimental results on \emph{val-2017}.
The evaluation result on VOC2007 \emph{test} in Table \ref{tab:voc} also shows the same trend even if the dataset is changed.

%%%%%%%%% Conclusion %%%%%%%%%

\section{Conclusion}
In this paper, we have presented a novel object detection model that reduces the computational cost by skipping the calculation on the background region with sparse convolution.
We use an OMG network to accomplish this purpose and integrate them in an end-to-end manner.
The experimental results with three popular network architectures and two datasets have shown that the proposed method can greatly reduce the computational cost while keeping comparable detection accuracy compared to the original models.
In addition, experiments with GT masks have shown that there is a lot of room for improvement in terms of both accuracy and speed.

For future work, we plan to evaluate the effect of state-of-the-art segmentation algorithms, which may improve object detection accuracy.

\clearpage
% ---- Bibliography ----
%
% BibTeX users should specify bibliography style 'splncs04'.
% References will then be sorted and formatted in the correct style.
%
\bibliographystyle{splncs04}
\bibliography{egbib}
\end{document}